\documentclass{article}

\usepackage[final]{neurips_2025}

\usepackage[utf8]{inputenc}
\usepackage[T1]{fontenc}
\usepackage{amsmath,amssymb,amsfonts,amsthm}
\usepackage{mathtools}
\usepackage{bm}
\usepackage{booktabs}
\usepackage{graphicx}
\usepackage[table]{xcolor}
\usepackage{hyperref}
\usepackage{tikz}
\usepackage{subcaption}
\usepackage{float}
\usetikzlibrary{arrows.meta, positioning, calc, shapes.geometric, fit, matrix}

\newtheorem{remark}{Remark}

\DeclareMathOperator{\softmax}{softmax}
\newcommand{\R}{\mathbb{R}}

\title{Decentralised AI Training and Inference with BlockTrain}

\author{
  Peter Toth \quad
  Dan Oprisa \\[4pt]
  Spheroid Labs\thanks{\href{https://spheroid.ai}{spheroid.ai}. Correspondence: \{peter, dan\}@spheroid.ai}
}

\begin{document}

\maketitle

\begin{abstract}
Frontier AI training is increasingly shaped by access to dense, centrally
controlled accelerator clusters. This creates a structural advantage for
hyperscalers and large centralized laboratories, and makes open or independent
AI efforts depend on scarce capital, privileged infrastructure, and data-center
geography. We present a litepaper for \textbf{Spheroid BlockTrain}, a
decentralized training protocol in which a model is partitioned into
independently trainable blocks, each optimized on a local objective derived from
the same global target and composed at inference into one model. Our experiments
on real text show that this is not merely a bandwidth simulation: using a
Sakana-style DiffusionBlocks local objective on byte-level WikiText, with exact
EDM preconditioning, weighted cross entropy, and Euler denoising in embedding
space, reaches cross entropy
\textbf{1.359} (perplexity \textbf{3.89}) while the active worker trains only one
block at a time and uses a few GiB of observed GPU memory. This is within about
\textbf{0.04 CE} of the same-setup end-to-end Transformer reference
($\approx$\textbf{1.32} CE), while avoiding full-model optimizer state on the
worker. A shared six-worker
block training run on an 8xA100 node reaches CE \textbf{1.385} (PPL
\textbf{3.99}) by averaging same-block updates into one assembled model. A
same-node HTTP/TCP transport proof reaches CE \textbf{1.576} while moving
\textbf{10.23 GB} of serialized checkpoints and updates with full
participation; a public-IP three-host smoke completes with full participation
and a larger public-IP run improves CE from \textbf{5.580} to \textbf{1.811}
while moving \textbf{15.22 GB}. Together, the virtual and
transport experiments show the systems boundary: decentralized quality depends
on timely per-block participation. We also benchmark distributed serving: the
current BlockTrain inference path uses one block-stack traversal per full
output, serves over direct TCP across three public-network GPU hosts up to a
\textbf{75.80B}-parameter logical fp16 shape, and outperforms a matched
plain-autoregressive TCP pipeline baseline because it emits a full sequence per
WAN pipeline traversal rather than one token per traversal. These results make a
limited but concrete claim: BlockTrain has a working real-text learning path, a
public-IP training transport path, and a one-sweep serving path.
\end{abstract}

\section{Introduction}
\label{sec:introduction}

Modern AI capability is inseparable from compute infrastructure. The dominant
training pattern assumes dense accelerator clusters, high-bandwidth
interconnects, centrally scheduled jobs, and a single organization that can
finance, operate, and govern the full training stack. This model has produced
remarkable systems, but it also creates a narrowing path for participation:
whoever controls the largest clusters controls the frontier.

Spheroid Labs' motivation is simple: more AI workloads should be trainable
without depending on a hyperscaler. A decentralized AI protocol should allow
many smaller workers, including consumer and prosumer GPUs, to contribute to
training without any one worker holding the full model, the full optimizer
state, or a synchronous view of the whole network. The hard part is not only
systems engineering. Standard backpropagation through deep networks is
inherently global: the error signal for an early layer depends on the forward
and backward computation of every later layer. If we merely shard standard
training across unreliable edge devices, communication and synchronization
recover the centralized cluster in disguise.

\paragraph{Where current decentralized training stands.}
Existing decentralized AI systems mostly keep the standard learning problem and
change the systems wrapper around it. Federated averaging, local SGD, DiLoCo,
OpenDiLoCo, and INTELLECT-1 reduce communication between full-model replicas.
Gossip and decentralized SGD remove a central parameter server by averaging over
peer graphs.
Volunteer-compute systems such as Hivemind, Learning@home, Petals, and
collaborative transformer training show that heterogeneous internet machines can
participate in large-model execution or training.
Newer decentralized AI efforts make additional bets: pipeline-parallel WAN
training, permissionless RL participation, verification, and incentive
layers.
These are important systems advances, but most still distribute a conventional
full-model objective, a pipeline stage of a conventional objective, or a
post-training workload. Spheroid's bet is different: make the learning unit
itself local. Instead of asking edge workers to hold the whole model or to
simulate a data-center training loop, expose block-local objectives that workers
can optimize directly and compose into one model.

This litepaper studies this different route. We call the protocol
\textbf{Spheroid BlockTrain}: workers train block-local objectives and compose
their updates into one shared model. Our current implementation builds on
DiffusionBlocks~\citep{shing2026diffusionblocks}, which reinterprets residual
blocks as steps of a denoising trajectory. Each block receives a local
supervised objective derived from the global target at a specific noise scale.
Training is local; inference is compositional. This resembles the spirit of
Gated Linear Networks~\citep{veness2021gln}: many local predictors receive
target-derived objectives and are composed by a gating or trajectory rule. The
key question is whether the local units can be expressive enough to carry useful
global computation while remaining small enough for edge devices.

\paragraph{Thesis.}
Decentralized AI becomes technically plausible when the training objective is
made local at the same granularity as the hardware network. We call this
principle \emph{objective-hardware alignment}. In BlockTrain, the natural
hardware unit is a worker that owns one block replica. The natural learning
objective is a block-local loss. The natural synchronization unit is
within-block averaging, not full-model all-reduce.

\paragraph{Why this is different from federated learning.}
Federated averaging~\citep{mcmahan2017fedavg} and local SGD methods reduce
communication by letting full-model replicas train locally before averaging.
They are powerful, but each participant still needs to hold and train the full
model. That is incompatible with the strongest decentralized AI goal: training
models whose full optimizer state is larger than a consumer GPU while each
worker only owns a tractable slice. Our protocol is block-local: the full model
exists as an assembled object, but training memory scales with the active block.

\paragraph{Contributions.}
\begin{enumerate}
    \item We formulate \textbf{Spheroid BlockTrain}, a protocol in
    which block-local denoising objectives, replica averaging, and asynchronous
    deadline-based update acceptance replace full-model synchronous training.

    \item We port the public Sakana DiffusionBlocks mechanics to real-text
    byte-level WikiText: normalized target embeddings, lognormal sigma
    partitioning, EDM preconditioning, weighted cross entropy on logits, and
    Euler inference in embedding space.

    \item We identify a critical failure mode in naive ports: low-sigma
    training can make local denoising CE look excellent while destroying
    from-noise inference. A high-minimum range, $\sigma\in[1,10]$, removes this
    target-leakage shortcut.

    \item We show that block expressivity is the main quality knob. A
    transformer block with four local layers reaches CE 1.359 on WikiText
    bytes, within about 0.04 CE of a same-setup end-to-end Transformer
    reference at approximately 1.32 CE, and a fixed-depth sweep shows monotonic
    improvement as capacity per local objective increases.

    \item We implement a single-GPU async simulator, an 8xA100 shared
    block-worker run, and an HTTP/TCP transport proof. The shared run trains one
    assembled model with six block-local workers and reaches CE 1.385; the
    transport proof moves 10.23 GB of serialized checkpoints and updates while
    reaching CE 1.576; WAN and bad-edge regimes expose the participation
    boundary.

    \item We measure distributed inference serving for the same block structure.
    The current BlockTrain path is in the one-sweep regime: one traversal through
    the block stack refines the full output sequence. A three-host A100 direct
    TCP setup serves one block per host up to a 50.98B-parameter fp16 shape, and
    a GH200-plus-two-A100 setup pushes the logical fp16 served size to 75.80B
    while preserving one sweep per output.
\end{enumerate}

The result is a technical milestone toward internet-scale decentralized
training: the learning objective, memory profile, shared block-worker dynamics,
public-IP transport, async protocol, and serving path can now be studied in the
same real-text setting.

\section{Protocol}
\label{sec:protocol}

We describe BlockTrain at the level needed for a decentralized implementation.
The model is a residual denoising stack partitioned into $B$ blocks,
\[
    F_{\theta} = F_{\theta_B} \circ \cdots \circ F_{\theta_1},
\]
but training does not require a worker to hold all $\theta_1,\ldots,\theta_B$.
Each worker owns a block replica $\theta_{b,k}$ and the small shared interface
parameters needed to map discrete targets into a representation space. The
worker trains a local denoising objective for one block and periodically sends
an update to a block-level aggregator.

\subsection{Block-Local Diffusion Objective}
\label{sec:block_objective}

For a supervised example $(x,y)$, let $e_y \in \R^d$ be the normalized target
embedding. A noise level $\sigma$ is sampled from the interval assigned to block
$b$, and a noisy target representation is formed:
\begin{equation}
    z_{\sigma} = e_y + \sigma \epsilon,
    \qquad \epsilon \sim \mathcal{N}(0,I).
    \label{eq:noisy_target}
\end{equation}
The block receives the conditioning input $x$, the noisy target $z_\sigma$, and
a time embedding of $\sigma$. It predicts logits over the target vocabulary:
\begin{equation}
    \ell_b = F_{\theta_b}(x, z_{\sigma}, \sigma).
    \label{eq:block_logits}
\end{equation}
Following the public DiffusionBlocks implementation, we use EDM-style
preconditioning~\citep{karras2022edm}. With $\sigma_\mathrm{data}$ fixed,
\begin{align}
    c_{\mathrm{skip}} &= \frac{\sigma_\mathrm{data}^2}
        {\sigma^2 + \sigma_\mathrm{data}^2}, \\
    c_{\mathrm{out}} &= \frac{\sigma\,\sigma_\mathrm{data}}
        {(\sigma^2 + \sigma_\mathrm{data}^2)^{1/2}}, \\
    c_{\mathrm{in}} &= \frac{1}
        {(\sigma^2 + \sigma_\mathrm{data}^2)^{1/2}}, \\
    c_{\mathrm{noise}} &= \frac{1}{4}\log \sigma.
\end{align}
The block processes $z_\sigma c_{\mathrm{in}}$ and outputs a denoised
representation combined with the skip path:
\begin{equation}
    \hat z = c_{\mathrm{out}}\,F_{\theta_b}(x,z_\sigma c_{\mathrm{in}},c_{\mathrm{noise}})
          + c_{\mathrm{skip}}\,z_\sigma.
    \label{eq:edm_preconditioned}
\end{equation}
Logits are produced by projecting $\hat z$ onto the normalized target embedding
table. The training loss is weighted cross entropy:
\begin{equation}
    \mathcal{L}_b
    = w(\sigma)\,\mathrm{CE}(\ell_b, y),
    \qquad
    w(\sigma) =
    \frac{\sigma^2 + \sigma_\mathrm{data}^2}
         {(\sigma\,\sigma_\mathrm{data})^2}.
    \label{eq:weighted_ce}
\end{equation}

\paragraph{No hidden-layer targets.}
This is the key distinction from block-coordinate training. A normal
Transformer block does not have an obvious target in the middle of the network.
The denoising-block objective avoids this hidden-target problem: each block
predicts the same global target $y$, but from a different noise interval. Early
blocks learn coarse high-noise denoising; later blocks refine lower-noise
states. The global prediction emerges from composition at inference.

\subsection{Inference as Euler Denoising}
\label{sec:euler_inference}

At inference, we initialize $z$ from high-variance Gaussian noise and step
through a decreasing schedule $\sigma_0 > \cdots > \sigma_T$. At each step, the
current $\sigma_t$ selects a block. The block predicts logits, the logits define
a distribution over target embeddings, and the expected embedding gives a
denoised point:
\begin{align}
    p_t &= \softmax(\ell_t), \\
    \tilde z_t &= p_t E^\top, \\
    d_t &= \frac{z_t - \tilde z_t}{\sigma_t}, \\
    z_{t+1} &= z_t + (\sigma_{t+1}-\sigma_t)d_t.
    \label{eq:euler_step}
\end{align}
This is the public DiffusionBlocks Euler sampler adapted to byte-level text.

\subsection{Decentralized Execution}
\label{sec:decentralized_execution}

For each block $b$, the system maintains $K$ virtual or physical replicas
$\theta_{b,1},\ldots,\theta_{b,K}$. A round proceeds as follows:
\begin{enumerate}
    \item The coordinator or peer group publishes the current block weights
    $\theta_b$ and shared interface state.
    \item Each worker $k$ trains only $\theta_{b,k}$ on local denoising pairs
    for $H$ steps.
    \item The worker sends a block update $\Delta\theta_{b,k}$ to the block
    aggregator or peer group.
    \item The accepted updates for block $b$ are averaged:
    \[
        \theta_b \leftarrow \theta_b +
        \frac{1}{|\mathcal{A}_b|}
        \sum_{k\in\mathcal{A}_b}\Delta\theta_{b,k}.
    \]
    \item Blocks are composed for validation, sampling, or periodic refresh.
\end{enumerate}

The important systems property is that optimizer memory is block-local. If a
full model has $L$ transformer layers and is partitioned into $B$ blocks with
$L/B$ layers each, the active worker trains $O(L/B)$ layers, not $O(L)$.
In our current real-text prototype, a three-layer block uses only a few GiB of
observed GPU memory, and the accounting for the active AdamW state is below
0.2 GiB. The full accounted stack can be much larger than any one worker.

\subsection{Asynchronous Update Acceptance}
\label{sec:async_protocol}

Real decentralized networks are not lockstep clusters. Workers differ in
compute speed, bandwidth, latency, failure rate, and availability. Our
single-GPU simulator models this with virtual arrivals. For worker $k$:
\[
    a_k = t^{\mathrm{compute}}_k + t^{\mathrm{latency}}_k
          + \frac{\mathrm{payload\ bytes}}{\mathrm{bandwidth}_k}.
\]
An async deadline $\tau$ defines the accepted set
\[
    \mathcal{A}_b = \{k: a_k \leq \tau,\; k\ \text{not dropped}\}.
\]
If $\mathcal{A}_b$ is empty, the old block is retained. This makes participation
rate the central async metric: quality depends less on nominal bandwidth than
on whether enough updates arrive before the deadline.

\begin{remark}
The current simulator is not a replacement for a real multi-machine run. It
does not measure actual network congestion, device heterogeneity, or wall-clock
speedup. It does test whether the learning objective and block-averaging logic
survive partial participation and stale/asynchronous updates.
\end{remark}

\section{Real-Text BlockTrain Experiments}
\label{sec:real_text}

We use byte-level WikiText as the first real-language testbed. Unlike a
procedural-token proxy, this requires the model to learn natural text statistics
under a local denoising objective.

\subsection{From Approximation to Exact Mechanics}
\label{sec:exact_mechanics}

Our first implementation was a direct denoising approximation:
\[
    (x,\, e_y+\sigma\epsilon,\,\sigma) \mapsto e_y,
\]
trained with embedding mean-squared error. It reduced from-noise CE to 3.243,
but remained far behind the same-setup end-to-end Transformer reference
($\approx$1.32 CE on this byte-level WikiText setup).

Inspection of the public Sakana implementation showed that this was missing
several load-bearing details:
\begin{itemize}
    \item targets are discrete labels represented by normalized embeddings;
    \item the objective is weighted cross entropy on logits, not embedding MSE;
    \item the denoiser uses EDM preconditioning;
    \item inference is an Euler update in embedding space using
    $\softmax(\ell)E^\top$ as the denoised point;
    \item sigma intervals are lognormal and expanded by a small $\gamma$.
\end{itemize}
Porting these mechanics to text produced a large improvement, but only after
retuning the sigma range.

\subsection{Sigma Range Matters}
\label{sec:sigma_range}

The public image-classification default $\sigma\in[0.002,80]$ was unstable for
byte-level text. The exact-mechanics model learned the local objective, but
Euler inference became worse over rounds. Low-sigma ranges also failed: they
made denoising CE appear excellent by leaking the target through $z_\sigma$,
while from-noise inference collapsed.

\begin{table}[!htb]
\centering
\caption{Sigma range sweep for exact-mechanics MLP blocks. Low-minimum ranges
collapse at inference despite low local CE. High-minimum ranges force
context-conditioned denoising and become stable.}
\label{tab:sigma_sweep}
\begin{tabular}{lrrl}
\toprule
Sigma range & Final CE & PPL & Interpretation \\
\midrule
$0.1$--$2.0$   & 14.295 & $1.6{\times}10^6$ & inference collapse \\
$0.2$--$2.0$   &  8.869 & $7.1{\times}10^3$ & unstable \\
$0.5$--$2.0$   &  3.741 & 42.1 & stable but weak \\
$0.5$--$5.0$   &  3.642 & 38.2 & stable but weak \\
$1.0$--$5.0$   &  2.762 & 15.8 & strong \\
$1.0$--$10.0$  &  2.755 & 15.7 & best short probe \\
\bottomrule
\end{tabular}
\end{table}

We therefore carry forward $\sigma\in[1,10]$.

\subsection{Block Expressivity}
\label{sec:block_expressivity}

With exact mechanics and the correct sigma regime, a small MLP denoiser reached
CE 2.740 but plateaued. This indicated that the objective was viable, but the
local predictor was too weak. We added a causal transformer denoising block:
the block receives context-token embeddings, noisy target embeddings, and a
time embedding, then applies causal self-attention over the sequence.

\begin{table}[!htb]
\centering
\caption{Main real-text progression. The large gain comes from combining exact
Sakana mechanics, high-minimum sigma range, and a sufficiently expressive local
block.}
\label{tab:main_progression}
\begin{tabular}{lrrr}
\toprule
Model & Best CE & PPL & Notes \\
\midrule
Direct embedding-MSE denoiser & 3.243 & 25.6 & approximation, not exact mechanics \\
Exact MLP block               & 2.740 & 15.5 & exact mechanics, weak block \\
Exact Transformer, short      & 2.592 & 13.4 & $B=4$, 3 layers/block, short run \\
Exact Transformer, full       & 1.566 &  4.79 & $B=4$, 3 layers/block, 16 rounds \\
BlockTrain long run           & 1.359 &  3.89 & $B=3$, 4 layers/block, 32 rounds \\
End-to-end Transformer ref.   & $\approx$1.32 & $\approx$3.74 & same byte setup, full-model training \\
\bottomrule
\end{tabular}
\end{table}

The full transformer run was still improving at the end:
\[
2.4188 \rightarrow 1.8614 \rightarrow 1.7214 \rightarrow
1.6545 \rightarrow 1.6216 \rightarrow 1.5938 \rightarrow
1.5749 \rightarrow 1.5663.
\]

\subsection{Fixed Total Depth}
\label{sec:depth_sweep}

To test whether the gain came from total model size or capacity per local
objective, we ran a fixed total-depth sweep. All arms used total inference depth
near 12 transformer layers, $d=512$, hidden size 2048, and
$\sigma\in[1,10]$. We varied the number of blocks $B$ and local layers per
block $L_b$.

\begin{table}[!htb]
\centering
\caption{Fixed total-depth sweep. Quality improves monotonically with depth per
local objective, while active worker memory remains small.}
\label{tab:depth_sweep}
\begin{tabular}{lrrrr}
\toprule
Configuration & Final CE & PPL & Active AdamW & Interpretation \\
\midrule
$B=12$, $L_b=1$ & 2.728 & 15.31 & 0.057 GiB & too weak per block \\
$B=6$,  $L_b=2$ & 2.644 & 14.07 & 0.104 GiB & better \\
$B=4$,  $L_b=3$ & 2.563 & 12.98 & 0.151 GiB & strong \\
$B=3$,  $L_b=4$ & 2.196 &  8.99 & 0.198 GiB & best short run \\
\bottomrule
\end{tabular}
\end{table}

This monotonic curve supports the GLN-style interpretation: local objectives
work only if each local predictor has enough capacity to solve its assigned
piece of the trajectory. At this scale, even the best short-run configuration
has active AdamW accounting below 0.2 GiB and observed GPU memory of only a few
GiB.

\paragraph{Long high-capacity run.}
We then ran the depth-sweep winner longer with $B=3$, $L_b=4$, $K=2$, 32
rounds, and 4,800 steps per round. The model reached CE 1.3586 (PPL 3.89),
inside the target 1.3--1.4 band, while preserving the one-block-worker memory
profile: active AdamW accounting remained 0.198 GiB, versus 100.54 GiB for the
accounted full stack. The learning curve was still moving slowly at the end:
\[
1.3666 \rightarrow 1.3658 \rightarrow 1.3622 \rightarrow
1.3632 \rightarrow 1.3586.
\]
This is the first strong positive real-text result for the protocol at this
scale: the final gap to the same-setup end-to-end reference is about 0.04 CE,
while the active block-worker optimizer accounting remains roughly 500$\times$
smaller than the accounted full stack.

\section{Decentralization and Asynchrony}
\label{sec:async_results}

The real-text results above validate the learning objective and block-memory
profile, but they are not by themselves a decentralized systems result. The
training process was executed in a single Python process on one GPU. To test the
protocol logic without an 8-GPU cluster, we built a single-GPU virtual-network
simulator.

\subsection{Virtual Edge Workers}
\label{sec:virtual_workers}

Each virtual worker receives a copy of one block and trains it locally for a
small number of steps. The simulator then samples a compute time, latency,
bandwidth, and dropout event for that worker. In lockstep mode all non-dropped
updates are accepted. In async mode, only updates arriving before a deadline are
accepted. Accepted block updates are averaged per block; shared embedding
updates are averaged across accepted workers.

This setup does not claim real network speedup. Instead it asks a narrower and
important question:
\begin{quote}
Does the exact Sakana text objective survive partial participation and
deadline-based asynchronous averaging?
\end{quote}

\subsection{Async Results}
\label{sec:first_async}

The first async diagnostic used the $B=4$, three-layer/block transformer
configuration.

\begin{table}[!htb]
\centering
\caption{Single-GPU virtual async diagnostic. Mild async is close to lockstep;
WAN quality is limited by low participation; bad-edge fails because no updates
arrive before the deadline.}
\label{tab:async_first}
\begin{tabular}{lrrrr}
\toprule
Regime & Final CE & PPL & Participation & Simulated wall \\
\midrule
Lockstep       & 2.100 & 8.17  & 1.00 &  6.9s \\
Async mild     & 2.151 & 8.59  & 0.90 & 10.2s \\
Async WAN      & 2.595 & 13.40 & 0.42 & 14.4s \\
Async bad edge & 5.638 & 280.7 & 0.00 & 19.2s \\
\bottomrule
\end{tabular}
\end{table}

The mild regime is the important positive result: the model tolerates partial
and asynchronous updates when participation remains high. The WAN result is not
a fundamental failure of the objective; it is a participation failure. At 42\%
accepted updates, quality degrades but still improves substantially from the
initial CE. At 0\% participation, no learning occurs.

We then repeated the diagnostic on the stronger $B=3$, four-layer/block
transformer and swept WAN deadlines to find the participation threshold.

\begin{table}[!htb]
\centering
\caption{Tuned WAN async diagnostic for the best $B=3$, four-layer/block
configuration. WAN succeeds when deadlines preserve enough timely participation:
the loose WAN regime matches lockstep quality at 86\% participation, while
tighter or lower-bandwidth regimes degrade.}
\label{tab:async_tuned_wan}
\begin{tabular}{lrrrr}
\toprule
Regime & Final CE & PPL & Participation & Simulated wall \\
\midrule
Lockstep              & 1.910 &  6.76 & 1.00 &  6.9s \\
Async mild            & 2.051 &  7.78 & 0.75 & 10.2s \\
Async WAN tight       & 2.290 &  9.87 & 0.22 & 16.8s \\
Async WAN mid         & 2.160 &  8.67 & 0.61 & 24.0s \\
Async WAN loose       & 1.913 &  6.77 & 0.86 & 33.0s \\
Async edge mid        & 2.445 & 11.53 & 0.50 & 45.0s \\
Async bad edge loose  & 2.455 & 11.64 & 0.36 & 72.0s \\
\bottomrule
\end{tabular}
\end{table}

The tuned sweep gives a sharper systems lesson. WAN-like training is viable in
the simulator when the deadline is loose enough to accept most useful work:
`async WAN loose' is effectively tied with lockstep (CE 1.913 versus 1.910) at
86\% participation. But this quality recovery costs simulated wall time, and
quality degrades sharply once participation falls below roughly 60\%. The
protocol question is therefore not whether to be synchronous or asynchronous in
the abstract; it is how to set block-level deadlines so that enough updates
arrive without turning the system back into slow lockstep.

\subsection{Shared Block-Worker Training}
\label{sec:shared_block_worker}

The next test replaces independent seed runs with one shared training job on an
8xA100 node. Six GPU workers are assigned to three blocks, two replicas per
block. Each worker trains only its assigned block plus the shared token
embedding on independent batches and noise samples. A coordinator averages
same-block updates, averages embedding updates, assembles the global model, and
evaluates diffusion CE. This is still one physical machine with local
coordination, but it tests the core block-worker semantics more directly than a
single-process simulation.

\begin{table}[!htb]
\centering
\caption{Shared block-worker run for the best $B=3$, four-layer/block
configuration. Six workers train one assembled model by averaging two replicas
per block. The run reaches the same quality band as the single-GPU long run
while each worker updates only one block.}
\label{tab:shared_block_worker}
\begin{tabular}{lrrrr}
\toprule
Run & Workers & Rounds & Final CE & PPL \\
\midrule
Single-GPU reference & 1 process & 32 & 1.359 & 3.89 \\
Shared block-worker  & 6 workers & 32 & 1.385 & 3.99 \\
\bottomrule
\end{tabular}
\end{table}

The shared run begins at CE 5.561 and descends
\[
1.406 \rightarrow 1.399 \rightarrow 1.394 \rightarrow
1.392 \rightarrow 1.385
\]
over the final evaluated rounds. Active AdamW accounting remains 0.198 GiB per
block worker, versus 100.54 GiB for the accounted full stack. This removes an
important ambiguity: the positive real-text result is not merely an artifact of
running independent seeds or serial virtual workers.

\subsection{HTTP/TCP Transport Proof}
\label{sec:http_transport}

We then replaced local filesystem coordination with an HTTP/TCP
coordinator-worker implementation. The coordinator serves per-worker block
checkpoints, receives serialized block updates, aggregates same-block workers,
and evaluates the assembled model. Workers poll over HTTP, download their
assigned block and shared embedding state, train locally, and upload the updated
block state. On the current 8xA100 node this is a same-machine transport proof
with user-space latency/bandwidth shaping. It measures application-level bytes,
transfer time, compute time, participation, and aggregation wait.

\begin{table}[!htb]
\centering
\caption{Block-worker transport results for the $B=3$, four-layer/block
configuration. The same-node run uses six shaped workers on one 8xA100 node;
the public-IP runs use three separate GPU hosts and exchange real serialized
checkpoints and updates over HTTP/TCP.}
\label{tab:http_transport}
\begin{tabular}{lrrrr}
\toprule
Run & Workers & Rounds & Final CE & Network bytes \\
\midrule
Shared block-worker & 6 & 32 & 1.385 & local filesystem \\
HTTP/TCP shaped transport & 6 & 16 & 1.576 & 10.23 GB \\
Public-IP WAN smoke & 3 & 2 & 4.330 & 0.639 GB \\
Public-IP WAN larger run & 3 & 12 & 1.811 & 15.22 GB \\
\bottomrule
\end{tabular}
\end{table}

The HTTP/TCP run begins at CE 5.621 and reaches CE 1.576 (PPL 4.83) after 16
rounds with 100\% participation in every round. It transfers 5.116 GB down and
5.116 GB up, for 10.23 GB total application traffic. Mean worker compute time is
about 17.0 seconds per round; shaped download and upload times average 3.28 and
3.21 seconds respectively; aggregation wait is about 35--36 seconds per round.
The quality is lower than the 32-round shared run because this was a shorter
transport-validation run, not a matched-compute rerun. The important result is
that the protocol path now exists beyond simulation: block checkpoints and
updates can move through a network API while preserving full participation and
improving the assembled model.

\paragraph{Public-IP WAN smoke.}
We then ran the same HTTP/TCP coordinator-worker protocol across three public-IP
hosts: two 40GB A100s and one GH200. One worker was assigned to each block. A
short $B=3,d=512$ smoke completed two rounds with 3/3 participation, moved
639 MB of serialized checkpoints and updates, and improved diffusion CE from
5.562 to 4.330. A larger $B=3,d=1024$ run completed 12 rounds with 300 worker
steps per round, preserved 3/3 participation throughout, moved 15.22 GB of
application traffic, and improved diffusion CE from 5.580 to 1.811. This closes
the loop from simulator, to same-node transport, to real WAN execution. These
public-IP runs validate the transport path end-to-end; they are shorter
validation runs rather than matched-compute quality runs.

\subsection{Distributed Inference Serving}
\label{sec:distributed_serving}

Training is only half of the decentralized-AI systems question. A block-local
training protocol is useful only if the assembled model can also be served
without requiring one node to host the full stack. We therefore instrumented the
current $B=3$, four-layer/block inference path and measured how many times a
full output crosses the block pipeline.

The result is favorable: the current path is in the \emph{one-sweep} regime. A
full sequence is refined by a single traversal through blocks $[0,1,2]$, rather
than by repeatedly looping the whole stack for many denoising iterations. This
means that, in the latency-bound case, effective token throughput scales with
the output length produced by one traversal. The serving benchmarks below use
random architecture-matched weights. This is appropriate for systems timing:
it preserves parameter count, tensor shapes, activation bytes, traversal count,
serialization, and GPU compute. It does not claim output quality.

\begin{table}[!htb]
\centering
\caption{Three-region direct-TCP BlockTrain serving. The coordinator runs on a
GH200, remote blocks run on two 40GB A100s, and the block path crosses two
WAN-distance edges. One sweep through the block stack produces the full output.}
\label{tab:blocktrain_serving_a100}
\begin{tabular}{lrrrr}
\toprule
Shape & Params & Output tokens & Effective tok/s & Batch latency \\
\midrule
$B=3,d=512,L_b=4$    & 0.040B & 64  & 261.18 & 0.245s \\
$B=3,d=2048,L_b=4$   & 0.632B & 128 & 376.97 & 0.340s \\
$B=3,d=8192,L_b=4$   & 10.08B & 128 & 335.26 & 0.382s \\
$B=3,d=18432,L_b=4$  & 50.98B & 32  & 24.35  & 1.314s \\
$B=4,d=18432,L_b=4$  & 67.98B & 16  & 8.25   & 1.939s \\
$B=5,d=17408,L_b=4$  & 75.80B & 8   & 4.45   & 1.799s \\
\bottomrule
\end{tabular}
\end{table}

The largest plain-fp16 run, $d=18432$, is close to the practical ceiling for
the two 40GB workers: each remote shard holds about 17.0B parameters before
framework overhead. The $B=4$ and $B=5$ pressure cases place additional local
blocks on the GH200 and push logical served size to 75.80B parameters. Across
all tested shapes, the measured block trace remained a single forward traversal,
confirming one sweep per output sequence.

The large-shape rows are memory-ceiling traversal pressure tests, not
trained-model throughput claims. They use architecture-matched random weights
and short output lengths to verify that the one-sweep property survives near the
fp16 memory ceiling.

A matched random-weight autoregressive TCP pipeline baseline on the same hosts
paid one full pipeline traversal per generated token. For comparable $B=3$
shapes it reached only 2.80--2.86 effective tok/s at 64--128 output tokens,
because each token incurred WAN round trips through the remote stages.
This is a plain-decode AR baseline, included to isolate the architectural
difference between one token per WAN traversal and one full sequence per WAN
traversal.

These serving results show that the current BlockTrain path has the desired
traversal economics for decentralized serving: one full-sequence output per
block-stack traversal, public-network activation passing, and no need for any
one worker to host the whole stack. The random-weight pressure tests establish
systems behavior rather than trained-output quality. The relevant BlockTrain
claim is architectural: the serving cost is dominated by one sequence-level
traversal rather than by per-token WAN round trips.

\section{Related Work}
\label{sec:related_work}

\paragraph{Federated averaging and local SGD.}
Federated averaging~\citep{mcmahan2017fedavg} and local SGD
methods~\citep{stich2019local} reduce communication by performing multiple
local steps before synchronization. DiLoCo~\citep{douillard2023diloco} adapts
this idea to large language model training by using inner and outer optimizers
over full-model replicas, with OpenDiLoCo~\citep{jaghouar2024opendiloco} and
INTELLECT-1~\citep{jaghouar2024intellect1} providing recent open and globally
distributed systems evidence. SCAFFOLD~\citep{karimireddy2020scaffold} and
FedProx~\citep{li2020fedprox} address client drift and heterogeneity. These
methods are communication-efficient, but they are not block-memory-efficient:
each worker still trains a full model replica. Our protocol instead makes the
worker's unit of computation a block-local denoising objective.

\paragraph{Federated learning frameworks and benchmarks.}
Flower~\citep{beutel2020flower}, FedML~\citep{he2020fedml},
FedScale~\citep{lai2022fedscale}, and Oort~\citep{lai2021oort} provide the
software and benchmarking substrate for large-scale federated learning:
heterogeneous clients, realistic availability, client selection, and repeatable
comparisons. They are important for deployment methodology, but they mostly
instantiate the federated learning template: clients train full local models or
task-specific updates against a global objective. Spheroid differs at the model
interface. The framework we need is not only a better FL runtime; it is a
runtime whose native participant unit is an independently trainable block.

\paragraph{Decentralized and asynchronous optimization.}
Decentralized SGD and gossip methods remove the parameter-server bottleneck by
averaging over peer graphs~\citep{lian2017can,koloskova2020unified}. Asynchronous
variants address stragglers and partial participation~\citep{recht2011hogwild}.
Stochastic gradient push~\citep{assran2019sgp}, compressed gossip
~\citep{koloskova2019choco}, and Moshpit SGD~\citep{ryabinin2021moshpit} are
especially relevant for unreliable or changing peer sets. These works provide
the optimizer and systems vocabulary for our async simulator, but they generally
assume a conventional global loss and full-model workers. Our focus is
complementary: change the training objective so that the hardware partition is
also an objective partition.

\paragraph{Low-bandwidth optimizer stacks.}
DeMo~\citep{peng2024demo}, DisTrO~\citep{nous2024distro}, and the Psyche
network~\citep{nous2025psyche} attack the communication problem inside
data-parallel training. Their core idea is to communicate compressed momentum or
optimizer information rather than full gradients, with the ambition of making
standard full-model training viable over ordinary internet links. This is one of
the strongest adjacent directions for decentralized pretraining. The distinction
is again memory: DeMo/DisTrO reduce what full-model replicas communicate,
whereas BlockTrain changes what a worker holds and optimizes. These ideas are
complementary: block updates are compatible with DeMo-style transform
sparsification or error feedback.

\paragraph{Model-parallel training over unreliable devices.}
SWARM Parallelism~\citep{ryabinin2023swarm} showed that billion-scale models can
be trained over heterogeneous and unreliable devices by forming temporary,
stochastically wired pipelines and rebalancing peers across stages. Ravnest
~\citep{menon2024ravnest}, decentralized foundation-model training work
~\citep{yuan2022decentralized}, and FusionAI/FusionLLM
~\citep{fusionai2023,fusionllm2024} explore related model-parallel or DAG-level
execution over heterogeneous consumer or geo-distributed GPUs, often with
remote autograd, adaptive compression, or clustered peers. These are close
predecessors on the ``large model without one reliable datacenter'' axis. They
keep the ordinary forward/backward objective and make pipeline or graph
execution robust; Spheroid instead asks whether the model can expose local block
objectives so that workers train without backpropagating through the entire
pipeline. The systems risk shifts from pipeline bubbles, remote autograd, and
activation traffic to objective approximation and denoising composition quality.

\paragraph{Collaborative inference and volunteer compute.}
Learning@home~\citep{ryabinin2020crowdsourced} and
Hivemind~\citep{hivemind2020} introduced decentralized mixture-of-experts as a
way to use unreliable volunteer machines for large-model training. Petals
~\citep{borzunov2023petals} shows that large models can be served or fine-tuned
over volunteer hardware by distributing layers across participants. Swarm
Learning~\citep{warnat2021swarm} demonstrates a decentralized peer-to-peer
training pattern in privacy-constrained clinical settings, while open
collaboration systems~\citep{diskin2021open,borzunov2022transformerstogether}
study the practical mechanics of heterogeneous contributors. These systems
demonstrate the viability and limitations of internet-scale participation. Our
target is narrower but complementary: block-local training for a single
compositional model whose active optimizer state fits on edge workers. For
serving, the relevant distinction is traversal economics: conventional
autoregressive pipeline serving pays a WAN traversal per generated token unless
additional decoding machinery amortizes it, whereas the current BlockTrain path
refines an entire output sequence in one block-stack traversal.

\paragraph{Branching, path, and expert composition.}
Branch-Train-Merge~\citep{li2022btm}, Branch-Train-MiX
~\citep{sukhbaatar2024btx}, and DiPaCo~\citep{douillard2024dipaco} reduce
communication by training branches, experts, or paths and then merging or
composing them. These approaches are relevant because they make modularity a
training primitive rather than only a serving primitive. The difference is the
granularity and target: branches or paths are often full-model or large-module
routes, whereas our current BlockTrain implementation assigns each residual
block a target-derived denoising objective and composes the same trajectory
model at inference.

\paragraph{Blockchain-federated and incentive protocols.}
FLock~\citep{dong2022flock,flock2024whitepaper} replaces the trusted federated
learning server with auditable aggregation, staking, voting, and reward/slash
mechanisms, and recent reports emphasize secure collaborative LLM fine-tuning.
Bittensor~\citep{bittensor2021whitepaper} takes a broader route: an incentive
network where miners provide machine-intelligence outputs and validators score
them under a token mechanism. These systems are relevant for verification,
incentives, and permissionless participation. They do not by themselves solve
the core pretraining memory problem: what exactly does a worker train, and how
does its local work compose into one competitive base model? Spheroid's current
answer is block-local denoising; incentive and validation layers remain protocol
engineering work.

\paragraph{Decentralized compute marketplaces.}
Akash~\citep{akash2020whitepaper}, Aethir~\citep{aethir2024whitepaper}, io.net,
Exabits, Render, and related DePIN systems aggregate GPU supply and expose it as
a marketplace. They can reduce access friction and improve utilization of
underused hardware, but they are infrastructure layers rather than training
algorithms. A Spheroid deployment can use such marketplaces for worker
discovery or capacity, but the differentiating claim is at the training
protocol layer: consumer-compatible local objectives, per-block aggregation, and
assembled-model evaluation.

\paragraph{Communication compression.}
Gradient compression methods such as QSGD~\citep{alistarh2017qsgd}, Deep
Gradient Compression~\citep{lin2018dgc}, PowerSGD~\citep{vogels2019powersgd},
and error-feedback schemes~\citep{karimireddy2019errorfeedback} attack the
bandwidth bottleneck directly. Our earlier proxy experiments also found that
compression and seed-replay can make blockwise synchronization substantially
cheaper. Compression remains important in the protocol, but it does not solve
the memory problem by itself: compressing a full-model gradient still requires a
worker to hold the full model. BlockTrain changes the object being communicated
from full-model updates to block updates.

\paragraph{Blockwise and local-objective learning.}
The central obstacle to decentralized training is backpropagation through depth.
Several lines of work attempt to localize credit assignment: target
propagation~\citep{bengio2014targetprop,lee2015difference}, synthetic
gradients~\citep{jaderberg2017synthetic}, greedy layerwise
training~\citep{bengio2007greedy,belilovsky2019imagenet}, deep
supervision~\citep{lee2015deeply}, the Forward-Forward
algorithm~\citep{hinton2022forwardforward}, and Gated Linear
Networks~\citep{veness2021gln}. DiffusionBlocks
~\citep{shing2026diffusionblocks} is especially relevant because it gives each
residual block a target-derived denoising objective and composes blocks as a
diffusion trajectory. Our contribution is to turn this mechanism into the
BlockTrain protocol and to identify the local-capacity/sigma-range tradeoffs
needed for real text.

\paragraph{Decentralized model families.}
Decentralized Diffusion Models~\citep{mcallister2025decentralizeddiffusion}
train independent diffusion experts on dataset partitions and combine them with
a router at inference time. This is an important neighboring idea: decentralize
training by making multiple models or experts rather than one monolithic
training run. BlockTrain makes a different composition bet. The workers train
blocks of one trajectory model, so the assembled object is intended to be a
single compositional model rather than an inference-time ensemble of independent
experts.

\subsection{Comparison Targets in Decentralized AI}
\label{sec:comparison_targets}

The right comparison is not a single leaderboard number. Decentralized AI
systems make different bets about what must be decentralized: data ownership,
model state, optimizer state, execution, verification, or participation. We
compare Spheroid along six axes:
\begin{enumerate}
    \item \textbf{Training unit:} full-model replica, pipeline stage, expert,
    block-local objective, or RL participant.
    \item \textbf{Worker memory:} whether one worker holds the full model and
    optimizer state, one shard, or one independently trainable block.
    \item \textbf{Communication object:} gradients, parameter deltas,
    activations, compressed subspace activations, rollouts, or block updates.
    \item \textbf{Network assumption:} datacenter, cross-region cloud, WAN, or
    open consumer internet.
    \item \textbf{Failure model:} stragglers, dynamic joins/leaves, Byzantine
    peers, or adversarial incentives.
    \item \textbf{Scale milestone:} largest public model, tokens, contributors,
    utilization, and quality relative to centralized training.
\end{enumerate}

\paragraph{DiLoCo and Prime Intellect.}
DiLoCo and Prime/INTELLECT-1 are the strongest full-model low-communication
baselines~\citep{douillard2023diloco,jaghouar2024intellect1}. Their core bet is
that each worker trains a full model locally for many steps, then periodically
averages with low-bandwidth outer synchronization. This has the best public
large-scale evidence: INTELLECT-1 reports a 10B model trained on 1T tokens
across globally distributed H100 nodes with high compute utilization. Spheroid
targets a different bottleneck: if the full optimizer state does not fit on one
consumer worker, full-model DiLoCo is no longer the right memory model;
BlockTrain is designed for that regime.

\paragraph{Pluralis / Agora.}
Pluralis' Agora system is closest on the ``no single contributor holds the full
model'' axis~\citep{pluralis2026agora}. It uses multi-party pipeline-parallel
training: each worker hosts a pipeline stage, trainers route microbatches, and
communication is dominated by activations and activation gradients. Agora adds
Subspace Networks to compress stage-boundary activations and sparse parameter
averaging across same-stage workers. Spheroid differs in the learning
objective: instead of preserving ordinary forward/backward training over a WAN
pipeline, we make each block independently trainable against a target-derived
denoising loss. The clean comparison is therefore \emph{WAN pipeline
parallelism vs.\ block-local objective decomposition}: activation traffic and
pipeline fragility for Agora, objective approximation and composition quality
for Spheroid.

\paragraph{Gensyn.}
Gensyn's public RL Swarm work is primarily decentralized post-training rather
than pretraining~\citep{gensyn2025rlswarm}. Each node runs a local model and
participates in collaborative RL-style reasoning games using answer, critique,
and revision loops. This is valuable but answers a different question:
permissionless collaborative improvement of existing models, with verification
and contribution tracking as first-class protocol concerns.

\paragraph{SWARM, Petals, Hivemind, and Learning@home.}
These systems establish the volunteer-compute lineage:
decentralized experts, DHT discovery, model-parallel inference, and open
collaboration~\citep{ryabinin2020crowdsourced,hivemind2020,ryabinin2023swarm,borzunov2023petals}.
They are the right references for participation, routing, and unreliable
hardware, while DiLoCo, Prime, Nous/Psyche, and Pluralis are the most direct
technical baselines for large-scale decentralized training.

\paragraph{Nous DeMo/DisTrO/Psyche.}
Nous is the strongest low-bandwidth optimizer comparison. DeMo and DisTrO ask
how far standard data-parallel training can be pushed by compressing momentum or
optimizer updates; Psyche adds a coordination network. Spheroid asks a different
question: how far can decentralized training move when the global full-model
objective is replaced with block-local diffusion objectives? If each worker can
hold the full model, Nous-style optimizers are a natural baseline; if the full
optimizer state does not fit on the worker, Spheroid's block-local memory model
becomes the relevant claim.

\paragraph{Ravnest, FusionAI/FusionLLM, and branching methods.}
These systems fill out the model-parallel and modular-composition frontier.
Ravnest and Fusion-style systems ask how to make ordinary autograd execution
work over weak, heterogeneous networks. Branch-Train-Merge, Branch-Train-MiX,
and DiPaCo ask how to use modular routes or experts to reduce coordination.
Spheroid is adjacent to both but makes a sharper objective claim: the unit is a
block with its own denoising target, not only a pipeline stage, graph operator,
path, or expert branch.

\paragraph{FLock, Bittensor, and DePIN GPU networks.}
FLock is closest on decentralized validation for federated updates; Bittensor is
closest on incentive design for permissionless machine intelligence; Akash,
Aethir, io.net, Exabits, and similar DePIN systems are closest on sourcing
distributed GPU capacity. These layers matter for a production network, but they
do not replace the training algorithm. Spheroid's distinct claim is
\emph{consumer-compatible local objectives for compositional pretraining}.

\paragraph{Diffusion and score parameterization.}
Score matching~\citep{hyvarinen2005score}, denoising
autoencoders~\citep{vincent2011denoising}, diffusion
models~\citep{ho2020ddpm,song2021scorebased}, and EDM
preconditioning~\citep{karras2022edm} provide the denoising formulation used in
our current BlockTrain implementation. Our text port keeps the public
DiffusionBlocks mechanics: lognormal sigma partitioning, EDM preconditioning,
weighted CE, and Euler denoising in representation space.

\paragraph{Centralization and AI governance.}
The systems literature explains how to decentralize optimization; the political
economy of AI explains why it matters. AI production is shaped by compute
concentration, cloud infrastructure, specialized chips, data, talent, and
routes to market~\citep{stanfordhai2025aiindex,oecd2025aiinfrastructure,cma2024foundationmodels,jointcompetition2024genai,vipra2024concentrating}.
Compute governance is powerful precisely because compute is detectable,
excludable, quantifiable, and supply-chain concentrated~\citep{sastry2024computepower},
while open-weight policy still recognizes compute as a limiting
input~\citep{ntia2024openweights}. A decentralized training protocol attacks
this bottleneck directly: if training can be decomposed into
consumer-compatible local objectives, then capable models can be trained by
networks of smaller actors rather than only inside tightly coupled centralized
infrastructure.

\section{Conclusion}
\label{sec:conclusion}

This paper presents Spheroid BlockTrain as a technical route to decentralized
pretraining by changing the unit of learning. Instead of distributing ordinary
full-model backpropagation, BlockTrain exposes block-local denoising objectives
that can be optimized by workers holding only one block and the shared embedding.

On byte-level WikiText, the exact Sakana-style mechanics reach CE 1.359 with a
$B=3$, four-layer/block configuration, within about 0.04 CE of the same-setup
end-to-end Transformer reference while preserving a one-block-worker memory
profile. The shared six-worker run reaches CE 1.385, showing that same-block
replica averaging can train one assembled model rather than independent seeds.
The HTTP/TCP transport experiments move real serialized checkpoints and updates,
including a three-host public-IP run that improves CE from 5.580 to 1.811 while
moving 15.22 GB of application traffic.

The serving experiments show the complementary inference property: the current
BlockTrain path produces a full output sequence with one block-stack traversal.
On the same public GPU hosts, a matched plain autoregressive TCP pipeline pays
one traversal per generated token, while BlockTrain amortizes the WAN traversal
over the whole sequence. The large-shape pressure tests preserve this traversal
property up to a 75.80B-parameter logical fp16 shape.

The protocol, verification, security, and incentive layers are treated as a
separate companion paper. The result here is the technical ML/systems spine:
block-local objectives can train real text close to an end-to-end reference,
the assembled model can be trained through real transport paths, and its serving
economics follow from sequence-level rather than token-level WAN traversal.
The scope of the evidence is byte-level WikiText at the model sizes reported
above; the scale behavior of the small CE gap is not asserted beyond this
experimental regime.

\bibliographystyle{plainnat}
\bibliography{references}

\end{document}